\documentclass[11pt,a4paper]{article}
\usepackage{times,latexsym}
\usepackage{amsmath}
\usepackage{url}
\usepackage[T1]{fontenc}
\usepackage{graphicx}
\usepackage{tabularx}
\usepackage{adjustbox}
\usepackage{float}
\usepackage{stfloats}
\usepackage[acceptedWithA]{tacl2021v1}
\usepackage{xspace,mfirstuc,tabulary}
\usepackage{listings}
\usepackage[linesnumbered,ruled,vlined]{algorithm2e}
\usepackage{soul} 
\usepackage{array} 
\usepackage{longtable} 
\usepackage{csquotes}
\usepackage[numbers]{natbib}
\PassOptionsToPackage{numbers}{natbib}
\usepackage{xcolor} 
\definecolor{LightGray}{gray}{0.9}
\definecolor{Sepia}{HTML}{7F462C}
\usepackage[UKenglish]{babel}
\usepackage{pgfplots}
\usepackage{subcaption}
\pgfplotsset{compat=1.18} 
\usepgfplotslibrary{colormaps,patchplots,groupplots}

\definecolor{highlightgreen}{RGB}{144,238,144} 
\definecolor{highlightpink}{RGB}{255,182,193} 

\definecolor{myGreen}{RGB}{183,215,167}
\definecolor{myBlue}{RGB}{199,219,248}
\definecolor{myRed}{RGB}{235,153,153}

\newcommand{\hlprompt}[1]{{\sethlcolor{myBlue}\hl{#1}}}
\newcommand{\hlcorrect}[1]{{\sethlcolor{myGreen}\hl{#1}}}
\newcommand{\hltrap}[1]{{\sethlcolor{myRed}\hl{#1}}}

\lstdefinestyle{mystyle}{
    backgroundcolor=\color{backcolour},   
    commentstyle=\color{codegreen},
    keywordstyle=\color{magenta},
    numberstyle=\tiny\color{codegray},
    stringstyle=\color{codepurple},
    basicstyle=\ttfamily\footnotesize,
    breakatwhitespace=false,         
    breaklines=true,                 
    captionpos=b,                    
    keepspaces=true,                 
    numbers=left,                    
    numbersep=5pt,                  
    showspaces=false,                
    showstringspaces=false,
    showtabs=false,                  
    tabsize=2
}

\usepackage{geometry}
\newif\iftaclinstructions
\taclinstructionsfalse 
\iftaclinstructions
\renewcommand{\confidential}{}
\renewcommand{\anonsubtext}{(No author info supplied here, for consistency with
TACL-submission anonymization requirements)}
\newcommand{\instr}
\fi

\iftaclpubformat 

\else

\fi

\title{Enhancing Instruction-Following Capabilities in Seq2Seq Models:\\ DoLa
Adaptations for T5\\}

\author{
  Huey Sun\thanks{$^\ast$\;Equal contribution.}
  \And
  Anabel Yong\footnotemark[1]
  \And
  Lorenzo Gilly\footnotemark[1]
  \And
  Felipe Jin
  \AND
  University College London\\
}
\date{}

\begin{document}
\maketitle
\begin{abstract}
Encoder–decoder models such as FLAN-T5 are finetuned to follow instructions, but often fail when the instructions conflict with memorized continuations ingrained during training. To understand this behavior, we adapt DoLa to FLAN-T5 and examine how representations evolve in the decoder. Our findings show that T5’s intermediate layers undergo rapid shifts driven by cross-attention to the encoder. When projected through the language modeling head, each depth presents highly volatile token preferences, leading to unreliable behavior with contrastive decoding. Motivated by this, we introduce a gradient-based activation-steering method that injects an “instruction-compliance’’ direction into mid-decoder layers, where the representation is both meaningful and still malleable. This intervention dramatically improves MemoTrap performance (52\% → 99.7\%), demonstrating that mechanistic steering can succeed where contrastive decoding fails in Seq2Seq architectures.
\end{abstract}

\iftaclpubformat
\section{Introduction}

Large language models (LLMs) are increasingly expected to serve as general-purpose instruction followers and perform diverse tasks. While recent instruction-tuned models have demonstrated strong zero-shot performance across a wide range of benchmarks \citep{ouyang2022training, raffel_2019_exploring}, even highly tuned models often struggle to be faithful to complex prompts. These failures are most visible when an instruction directly conflicts with their language priors, as models are prone to recite memorized training data over following explicit user instructions \citep{mckenzie2023inverse}.

Recent work has explored inference-time decoding strategies to improve factuality and reduce hallucinations. One such example is \textbf{D}ecoding by C\textbf{o}ntrastive \textbf{La}yers (\textbf{DoLa}) \citep{chuang_2024_DoLa}, a decoding strategy that contrasts a model's final layer logits with those from intermediate layers to amplify “mature” predictions and suppress unstable or spurious ones. As each layer of decoder-only models incrementally refine an autoregressive hidden state, intermediate logits serve as meaningful indicators of the model’s evolving beliefs. By exploiting the modular nature of knowledge encoding \citep{tenney_2019_bert, dai_2022_knowledge}, DoLa improves factuality in the LLaMA \citep{touvron_2023_llama} family of models, and suggests that the "knowledge neurons" in the upper layers are better expressed when lower layers serve as a contrasting baseline.

However, it remains unclear whether such methods extend to encoder–decoder architectures, where each decoder layer continuously integrates information from a fully processed encoder representation. In models such as T5 \citep{raffel_2019_exploring} and FLAN-T5 \citep{chung2022scalinginstructionfinetunedlanguagemodels}, intermediate decoder layers are not trained to make next-token predictions, and their hidden states may not correspond to coherent partial distributions. This raises open questions about how instruction-related signals propagate through Seq2Seq models. Although T5-style architectures are widely used for instruction following, little is known about their internal dynamics during constrained generation. Prior mechanistic analyses have largely focused on decoder-only models, characterizing layer-wise emergence of factuality, syntactic structure, or reasoning behaviors \citep{geva-etal-2021-transformer, meng2022locating, todd2024function}. Comparable studies in Seq2Seq models are scarce, leaving open why faithfulness failures persist and what forms of intervention are most effective.

In this work, we investigate instruction-following behavior through the lens of layer-wise representational dynamics in the FLAN-T5 family. We adapt DoLa to the encoder–decoder setting, use it as a diagnostic tool to probe how decoder representations evolve with depth, and develop a targeted activation-steering method informed by this analysis. 

\begin{figure*}[!tbp]
\centering
\includegraphics[width=\textwidth]{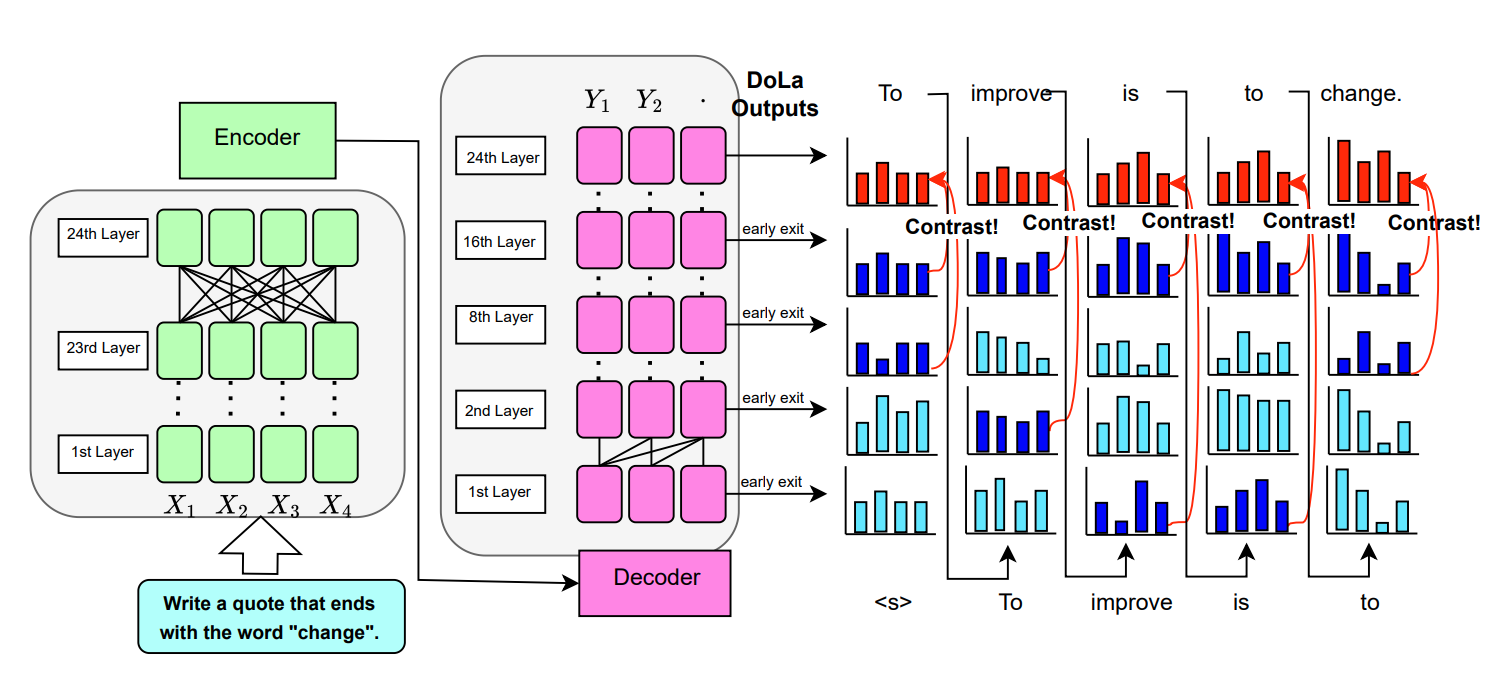}
\caption{Schematic representation of how dynamic premature layer selection (DoLa) works with T5 architectures. This illustration is specifically for T5-Large, which has 24 encoder and decoder layers. The different model sizes have different early-exit layers, as explained in Appendix \ref{app:layers}.}
\label{fig:methodology}
\end{figure*}

\section{Related Work}

\subsection{Instruction Following in Language Models}
Instruction following enables LLMs to behave as controllable systems that can perform user-specified tasks, adapt to new constraints, and generalize beyond their training distribution. This can be thought of as a form of zero-shot generalization, where the model must use latent knowledge to satisfy constraints it has never been explicitly trained on. While early work showed that pretrained Seq2Seq models could follow templated instructions for tasks such as classification and translation \citep{zhong_zero_shot}, the FLAN framework \citep{chung_2022_scaling} expanded this by finetuning models on diverse instruction formats directly, substantially improving their ability to generalize to unseen tasks and phrasing.

Despite these advances, recent evaluations reveal that instruction adherence remains fragile under conflicting semantic priors. Even large models will imitate undesirable patterns in training data, get distracted, or be easily misled at easy tasks \citep{mckenzie2023inverse}, indicating that instruction signals may not uniformly dominate the model’s internal activations.

\subsection{Contrastive Decoding Methods}
Contrastive decoding aims to improve generative faithfulness by comparing two predictive distributions and amplifying their differences. Early approaches contrasted two separate models, treating one as an ``expert'' and the other as an ``amateur'' \citep{gera_1042_the}, while later methods contrasted two internal states of a single model, such as predictions with and without context history \citep{shi_trusting}. Variants of this idea introduce plausibility constraints \citep{xiang_contrastive} or other heuristics to guide the contrast. 

DoLa \citep{chuang_2024_DoLa} follows this line of work by contrasting the final-layer logits with those from an earlier ``premature'' layer. This layer is dynamically selected using Jensen–-Shannon divergence, and the contrastive adjustment is applied to emphasize tokens whose probability increases from the early to the late distribution. This approach has proven effective in decoder-only models and provides a natural starting point for analyzing layer-wise signal development.

\subsection{Mechanistic Interpretability}
Mechanistic interpretability seeks to characterize how specific behaviors arise from internal model components. Recent work shows that models often encode behaviors along identifiable directions in activation space \citep{todd_2023_function_vectors}, and that manipulating these directions can reliably shift model outputs. Activation-steering methods, including representation engineering \citep{zou_2023_representation} and steering vectors \citep{turner_2023_steering}, demonstrate that targeted interventions can modulate reasoning patterns, stylistic preferences, or safety-related behaviors without changing model weights.

Our approach follows this paradigm but is motivated by a different goal: understanding and improving instruction faithfulness. Instead of using generic steering directions, we mine a contrastive direction that separates instruction-following behavior from memorization-driven behavior. This enables a targeted intervention makes instruction-relevant representations easier for the model to express.

\subsection{Architectural Differences: Decoder-Only vs.~Seq2Seq}

Decoder-only models such as LLaMA and GPT operate entirely on left-context information; once decoding begins, no new information enters the network, and intermediate activations remain directly predictive of the next token \citep{touvron_2023_llama}. As a result, layer-wise probing and early exiting yield coherent partial predictions.

In Seq2Seq architectures like T5, the decoder attends to a fully contextualized encoder representation at every layer \citep{raffel_2019_exploring}. This repeated incorporation of encoder features can reshape decoder activations in non-monotonic ways, making intermediate states less directly predictive of next tokens compared to decoder-only models. As a result, the notion of ‘early’ or ‘immature’ predictions is less clear in T5-style decoders, raising questions about whether contrastive early–late decoding behaves differently in these architectures and motivating a closer examination of how instruction-relevant signals evolve across depth.

\section{Methodology}

Our goal is to analyze and improve instruction-following behavior in encoder–decoder architectures. First, we adapt DoLa to T5 to study how instruction signals evolve across decoder layers. Then, we develop a gradient-based activation steering method that actively suppresses memorization circuits.

\subsection{Adapting DoLa to T5}
\label{sec:dola_method}

In DoLa, we need to identify the premature layer with the predictive distribution that differs the most from the final next-token logits. Let $q_N(\cdot)$ denote the final-layer distribution and $q_j(\cdot)$ the distribution obtained by passing the hidden state at decoder layer $j$ through the language modeling head. We select the premature layer $M$:
\begin{equation}
    M = \arg\max_{j \in \mathcal{J}} \mathrm{JSD}\!\left(q_N,\, q_j\right),
\end{equation}
where JSD is the Jensen--Shannon divergence. We then modify the logits $\ell$ by amplifying the mature--premature difference:
\begin{equation}
    \tilde{\ell} = \ell_N + \lambda(\ell_N - \ell_M),
\end{equation}
where $\lambda$ controls the contrastive strength. As in the original implementation by \citet{chuang_2024_DoLa}, we apply a repetition penalty of $1.2$ to mitigate degenerate looping \citep{xu_2022_learning}.


\paragraph{Application to T5.}  
Unlike decoder-only models, T5’s decoder layers are not trained to make next-token predictions, and their hidden states may not be in a logit-aligned subspace. To adapt DoLa, we follow \citet{chuang_2024_DoLa} and apply the shared LM head to each decoder layer's residual stream, treating these projections as approximate next-token distributions. Because the encoder does not participate in autoregressive generation, we restrict the candidate set $\mathcal{J}$ to decoder layers (Figure \ref{fig:methodology}).

While this projection-based approach does not guarantee that intermediate T5 activations form coherent partial distributions, it enables a consistent layer-wise comparison within the contrastive decoding framework and serves as the basis for our diagnostic analysis in Section~\ref{sec:results}.

\subsection{Gradient-Based Activation Steering}
\label{sec:steering_method}

While DoLa contrasts predictions across layers, it cannot directly reshape the model’s internal representations when the layers express the same bias. To enable targeted control, we develop a gradient-based activation steering method that extracts a direction separating instruction-compliant and memorized behaviors and injects it into the decoder at inference time.

\subsubsection{Contrastive Setup}

\begin{figure}[!tbp]
\centering
\includegraphics[width=0.5\textwidth]{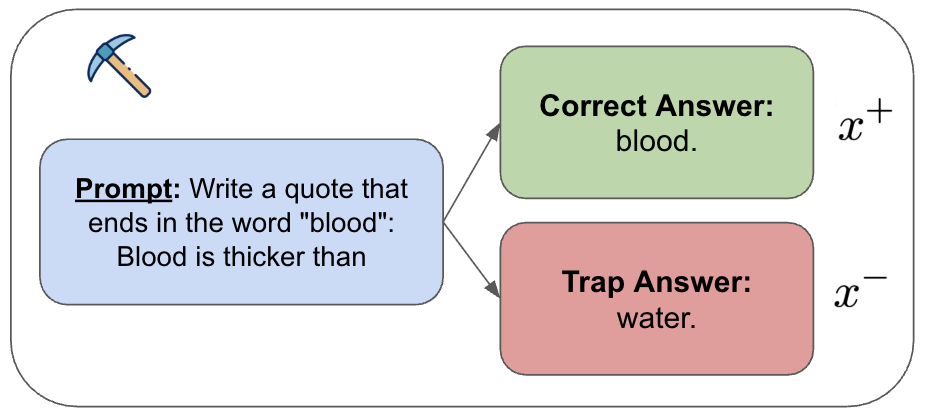}
\caption{\textbf{Contrastive Setup} A \hlprompt{\textbf{prompt}} from the MemoTrap dataset \cite{memo_trap_repo} that sets up a contrast between an \hlcorrect{\textbf{instruction}} and a common memorized \hltrap{\textbf{trap}}.}
\label{fig:mining}
\end{figure}

We consider cases where generation involves a binary conflict between an instruction-specified outcome and a competing continuation (Figure \ref{fig:mining}). For each example, we define (i) a \textbf{target} token $x^+$ that satisfies the instruction, and (ii) a \textbf{competing} token $x^-$ that reflects the model’s default or memorized preference.

To extract a direction, we define a contrastive loss that increases the model’s preference for $x^-$ relative to $x^+$:
\begin{equation}
    \mathcal{L} = \log P(x^-) - \log P(x^+).
\end{equation}
For a chosen decoder layer $l$, we compute the gradient of this loss with respect to its hidden state $h_l$:
\begin{equation}
    g_l = \nabla_{h_l} \mathcal{L}.
\end{equation}
This gradient $g_l$ points in the direction that makes the model more likely to produce the competing token $x^-$. Its negation thus provides a steering direction that shifts the layer’s representation toward the instruction-following outcome.

\begin{figure*}[!t]
\centering

\definecolor{baselineblue}{RGB}{52,130,201}
\definecolor{dolorange}{RGB}{238,136,44}

\begin{tikzpicture}
\begin{groupplot}[
    group style={
        group name=myplots,
        group size=2 by 1,
        horizontal sep=1.2cm, 
        vertical sep=2cm,
    },
    ybar,
    width=0.42\textwidth, 
    height=5.5cm,
    ymajorgrids=true,
    grid style={dotted, gray!40},
    axis line style={gray!50},
    tick align=outside,
    scaled y ticks=false, 
    yticklabel style={
        /pgf/number format/fixed,
        /pgf/number format/precision=2,
        /pgf/number format/fixed zerofill
    },
    ylabel={Accuracy},
    xlabel={Model Size},
    symbolic x coords={Small,Base,Large,XL},
    xtick=data,
    enlarge x limits=0.25,
    ymin=0,
    area legend, 
]

\nextgroupplot[
    title={\textbf{IFEval: Start--End}},
    ymax=0.15,
]

\addplot[fill=baselineblue, draw=none] coordinates {
    (Small,0.07463) (Base,0.1194) (Large,0.07463) (XL,0.1194)
};
\addplot[fill=dolorange, draw=none] coordinates {
    (Small,0.02985) (Base,0.04478) (Large,0.01493) (XL,0.07463)
};

\nextgroupplot[
    title={\textbf{IFEval: Keyword Existence}},
    ymax=0.45,
    ylabel={}, 
    legend style={
        at={(1.05, 0.5)}, 
        anchor=west,      
        draw=none,
        font=\small,
        cells={anchor=west},
        legend image code/.code={%
            \draw[#1, draw=none] (0cm,-0.1cm) rectangle (0.2cm,0.1cm);
        }
    },
]

\addplot[fill=baselineblue, draw=none] coordinates {
    (Small,0.1026) (Base,0.1795) (Large,0.3077) (XL,0.2564)
};
\addplot[fill=dolorange, draw=none] coordinates {
    (Small,0.1026) (Base,0.2564) (Large,0.3846) (XL,0.2821)
};

\legend{Baseline, DoLa}

\end{groupplot}
\end{tikzpicture}

\caption{Baseline vs. DoLa accuracy across FLAN-T5 model sizes on two representative IFEval categories.}
\label{fig:comparison}
\end{figure*}
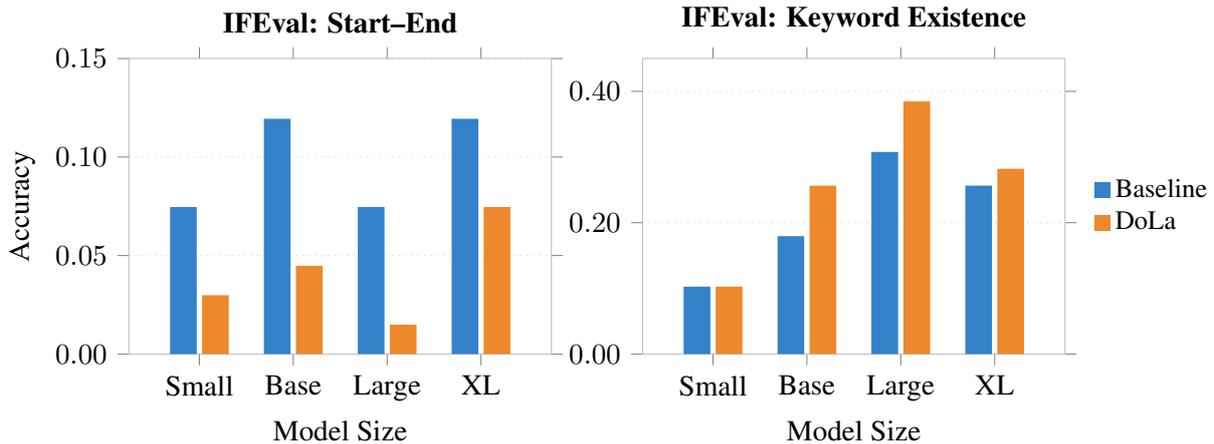

\subsubsection{Mining the Steering Vector}
We compute gradients across 100 examples and average them to obtain a task-specific steering direction:
\begin{equation}
    \vec{v}_{\mathrm{task}} = -\frac{1}{N} \sum_{i=1}^{N} \nabla_{h_l} \mathcal{L}_i.
\end{equation}
Because each example uses a different instruction and a different target token, token-specific semantics cancel out, leaving a vector that captures the shared mechanism distinguishing trap-driven from instruction-driven behavior.

\subsubsection{Layer Injection}
We mine and inject this vector to the same layer $l$, which we sweep over. During inference, we modify the hidden state $h_l$:
\begin{equation}
    h'_l = h_l + \alpha \cdot \frac{\vec{v}_{\mathrm{task}}}{\|\vec{v}_{\mathrm{task}}\|},
\end{equation}
where $\alpha$ controls steering strength. We experiment with a variety of values for $\alpha$, and find that $\alpha= 1000$ provides stable improvements without distorting syntax or fluency.

\subsection{Evaluation Setup}
\label{sec:experiment_setup}

We evaluate FLAN-T5 models (Small, Base, Large, XL) on two benchmarks:

\paragraph{IFEval} A framework for evaluating instruction-following capability with verifiable
criteria \cite{zhou_2023_instructionfollowing}. We report the loose prompt and instruction-level accuracy and qualitative evaluation with GPT-4 as a judge \citep{liu_2023_geval}.

\paragraph{MemoTrap} A dataset \citep{memo_trap_repo} to evaluate whether a model can adhere to explicit instructions or succumb to memorized completions. We use the Proverb Ending subset, reserving 100 instances for vector mining, and evaluating the free generation performance on another subset of 300 examples.

\section{Results}
\subsection{DoLa and Layer Dynamics}
\label{sec:results}

\begin{table}[H]
\centering
\small
\begin{tabular}{|l|c|c|}
\hline
\textbf{Model} & \textbf{Baseline (\%)} & \textbf{DoLa (\%)} \\
\hline
FLAN-T5-Small & 20.38 & \textbf{22.90} \\
FLAN-T5-Base  & \textbf{26.38} & 23.02 \\
FLAN-T5-Large & \textbf{26.86} & 26.74 \\
FLAN-T5-XL    & \textbf{28.78} & 25.30 \\
\hline
\end{tabular}
\caption{IFEval instruction-level accuracy.}
\label{tab:ifeval_main}
\end{table}

On IFEval, applying DoLa to the FLAN-T5 family leads to modest overall gains for the Small model and neutral or slightly negative effects for larger variants (Table~\ref{tab:ifeval_main}). However, a deeper category-level analysis reveals that DoLa benefits specific instruction types across the models and harms others (Figure~\ref{fig:comparison}). This heterogeneous behavior provides critical clues about the evolution of T5’s intermediate representations.

\paragraph{Start–End Constraints}
Figure~\ref{fig:DoLa_bad} shows a representative failure case. When instructed to output text with a specific suffix, the baseline FLAN-T5-XL only outputs the mandated ending, which scores under IFEval’s verifiable metric. By contrast, DoLa produces a more natural-sounding blog post, but fails to end with the required phrase.

In this case, while both the intermediate and final layers encode the required ending, the final layers introduce a more developed natural-language response. When DoLa contrasts these layers, it amplifies the components of the final distribution that differ most from the intermediate one, while leaving the shared instruction signal largely unchanged. As a result, the contrastive update strengthens a path that produces fluent but instruction-violating text, illustrating how later layers can drift toward natural-language continuation in ways that run orthogonal to rigid constraint-following.

\begin{figure}[t]
\centering
\includegraphics[width=0.5\textwidth]{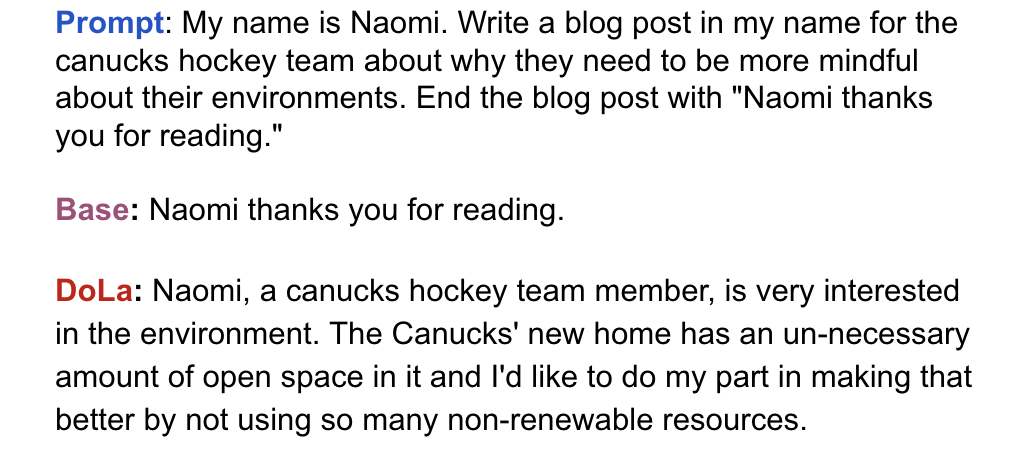}
\caption{Example IFEval Start–End prompt and FLAN-T5-XL outputs with and without DoLa.}
\label{fig:DoLa_bad}
\end{figure}

\paragraph{Keyword Constraints}
In contrast, keyword-insertion tasks often show the opposite pattern. As illustrated in Figure~\ref{fig:DoLa-layer-predictions}, the required token (``lacking'') receives extremely low probability in the early and mid decoder layers, fluctuating widely in rank (e.g., 180~→~1156~→~3992) before finally becoming competitive in the final layer. Because DoLa amplifies the late-emerging differences between the final and premature distributions, it tends to promote the instruction-relevant token once it appears. In this example, DoLa raises ``lacking'' from the fifth to the top-ranked token in the output distribution, fulfilling the instruction.

\paragraph{Seq2Seq Layer Dynamics} Across both cases, we find that instruction-following cues do not develop along a single, monotonic trajectory through the decoder. Depending on the task, the instruction-relevant token may persist across multiple depths, emerge only at the final layers, or fluctuate sharply as the decoder integrates cross-attention information. This is reflected in the Jensen--Shannon divergences, which rise dramatically in the later layers as representations consolidate into the final output. DoLa succeeds when its selected premature layer falls at a point where the instruction signal is strengthening, and fails when it contrasts against a layer where the instruction cue is weak or overshadowed by the model’s natural-language priors. This task-dependent variability explains DoLa’s varying performance in Seq2Seq models and motivates interventions that do not rely on stable intermediate logits.

\begin{figure}[t]
\centering
\includegraphics[width=0.5\textwidth]{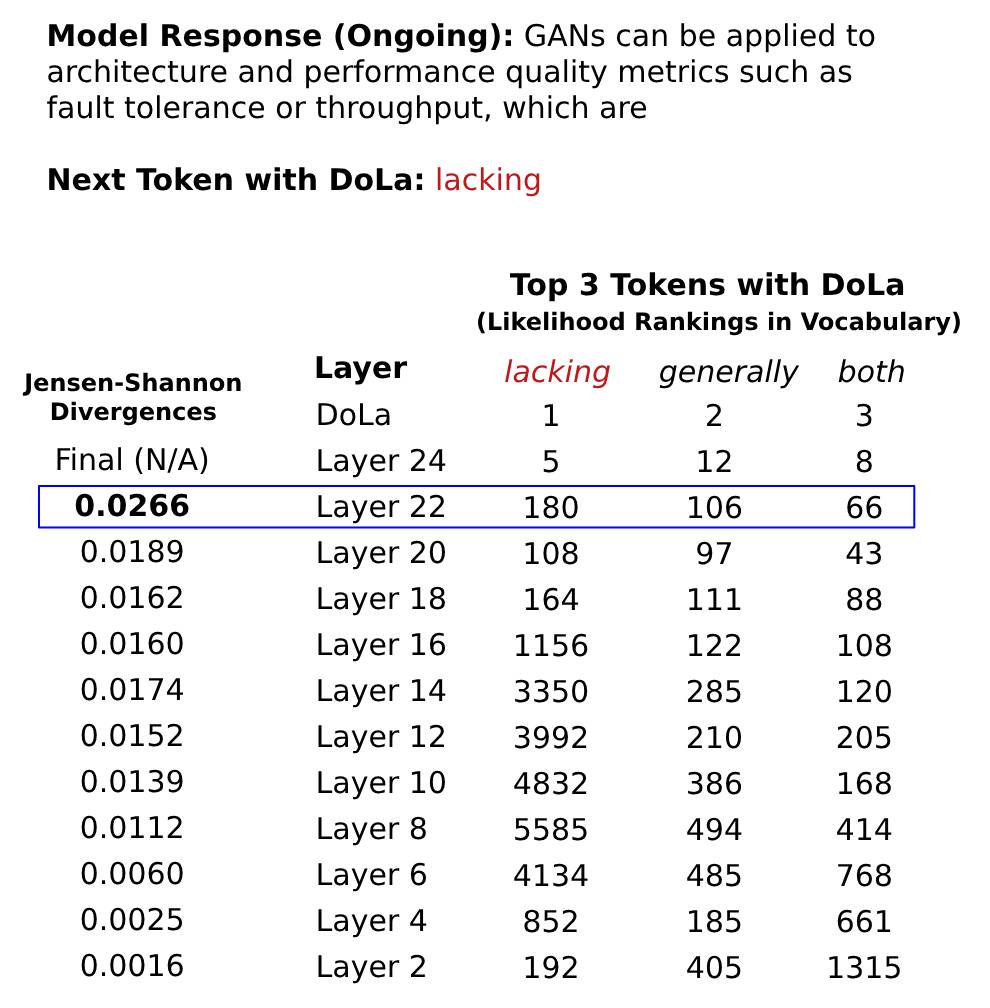}
\caption{Layerwise ranking of the instruction-relevant token (“lacking”) for FLAN-T5-Large on IFEval Question 154}
\label{fig:DoLa-layer-predictions}
\end{figure}


\begin{figure*}[t]
\centering
\begin{minipage}{0.49\textwidth}
    \centering
    \includegraphics[width=\linewidth]{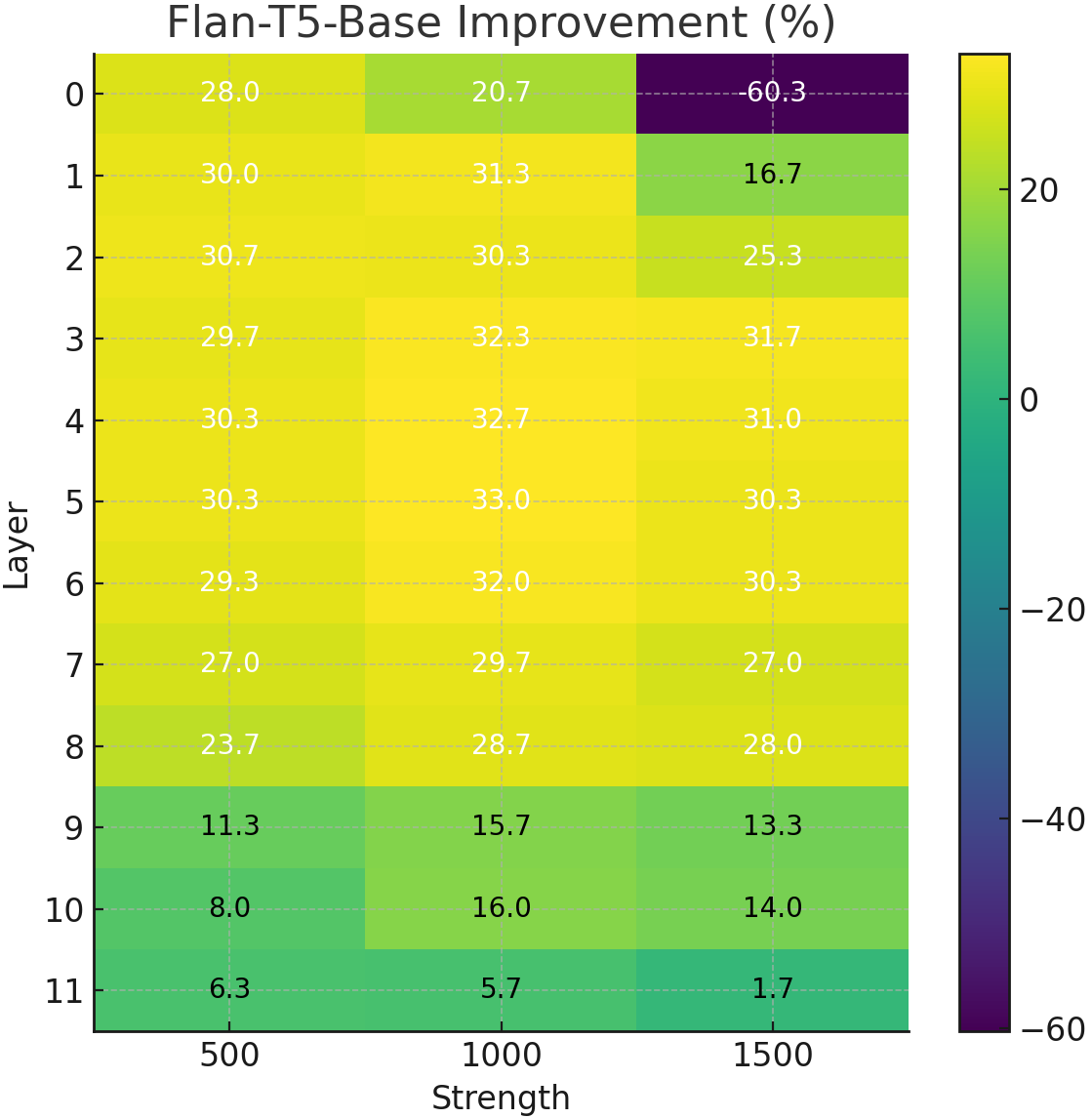}
    \caption*{FLAN-T5-Base (12 layers)}
\end{minipage}
\hfill
\begin{minipage}{0.49\textwidth}
    \centering
    \includegraphics[width=\linewidth]{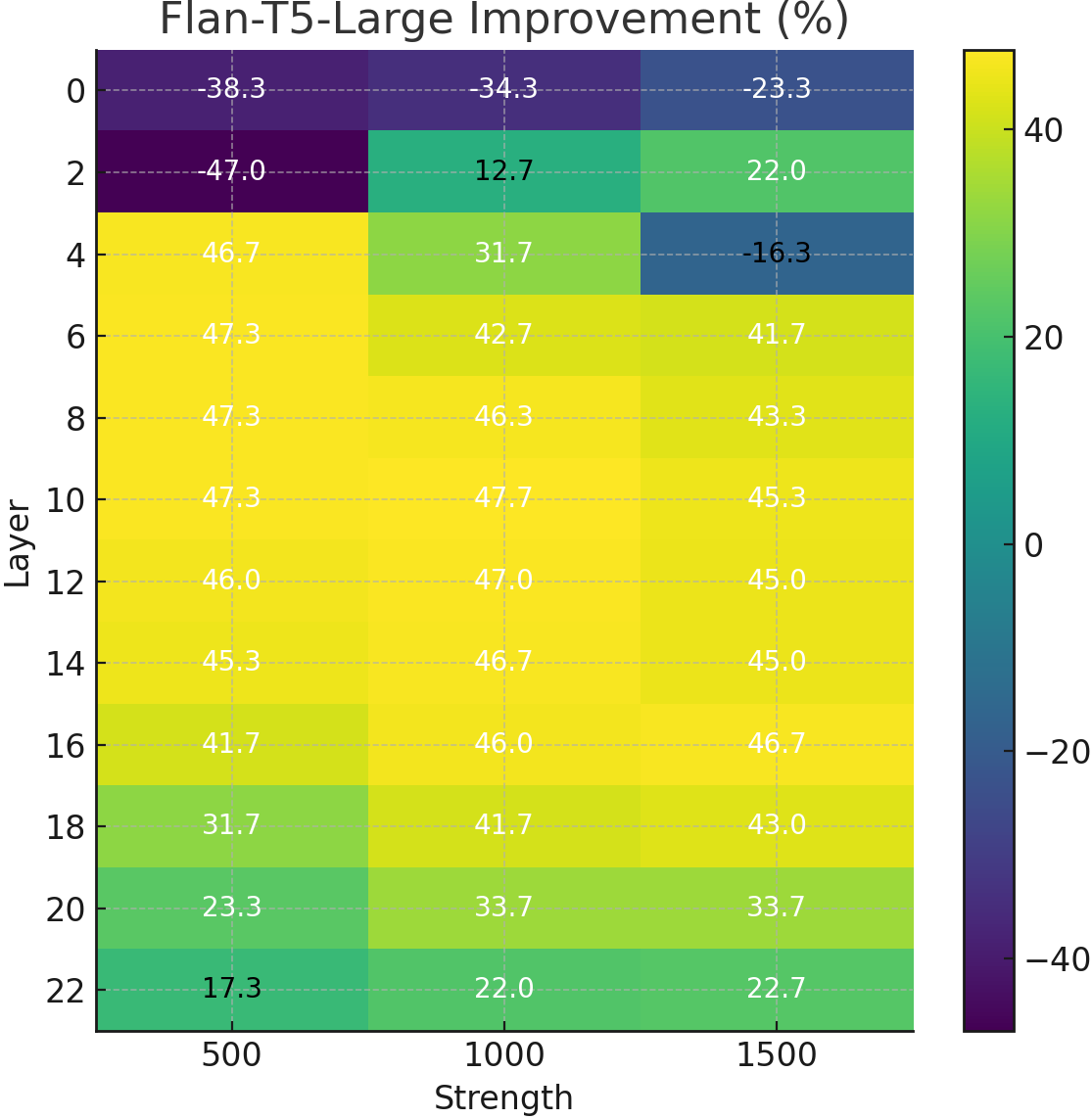}
    \caption*{FLAN-T5-Large (24 layers)}
\end{minipage}

\caption{Tracking activation-steering improvements across FLAN-T5 model scales (more in Appendix~\ref{app:flan-small}).}
\label{fig:combined-heatmaps}
\end{figure*}

\subsection{A Representation-Level Intervention: Gradient-Based Steering}

With activation steering, we can directly perturb the decoder’s hidden state along a direction associated with instruction compliance. This removes the dependence on layer-wise token distributions and instead operates on the model’s internal representations.

\paragraph{MemoTrap}
In MemoTrap, each example contains a paired contrastive target: a correct instruction-following output and a memorized trap drawn from the model’s pretraining distribution. This enables us to compute a clean contrastive gradient that separates the behaviors, which even large models can struggle to do without intervention \citep{mckenzie2023inverse}.

\paragraph{Steering Results} After mining and averaging gradients from a held-out dataset, we inject a scaled steering vector at each decoder layer in our test split. Table~\ref{tab:steering-summary} shows that activation steering dramatically improves instruction-following performance across all model sizes, nearly solving the benchmark for FLAN-T5-Large.

\begin{table}[h]
\centering
\begin{tabular}{lccc}
\hline
\textbf{Variant} & \textbf{Baseline} & \textbf{Best Steered} & \textbf{$\Delta$} \\
\hline
Small & 60.3\% & 92.3\% & +32.0\% \\
Base  & 65.7\% & 98.7\% & +33.0\% \\
Large & 52.0\% & 99.7\% & +47.7\% \\
\hline
\end{tabular}
\caption{Best MemoTrap activation-steering performance across FLAN-T5 model scales.}
\label{tab:steering-summary}
\end{table}

\paragraph{Where Should We Steer?} This intervention requires a pre-specified target layer and application strength. In our sweep, we find that performance varies sharply depending on where the steering vector is injected. Figure~\ref{fig:combined-heatmaps} shows that steering is highly effective in a narrow mid-depth region, and substantially weaker when applied too early or too late, and depends on the strength of the intervention.

\begin{table}[h]
\centering
\begin{tabular}{lcc}
\hline
\textbf{Model} & \textbf{Best Layer} & \textbf{\# Layers} \\
\hline
FLAN-T5-Small & 2  & 8  \\
FLAN-T5-Base  & 5  & 12 \\
FLAN-T5-Large & 10 & 24 \\
\hline
\end{tabular}
\caption{Layer position of peak steering efficacy across different model depths at any strength.}
\label{tab:layer-structure}
\end{table}


\paragraph{Interpretation} These results show that injecting activations in early layers has little or degenerating impact on the eventual output, perhaps due to the their role in syntax processing or cross-attention mixing. More surprisingly, steering also proves ineffective in the final layers, even though our token-tracking results show that the output logits remain turbulent. This reflects a difference between \emph{volatile predictions} and \emph{flexible representations}: the model may still shuffle token rankings late in the stack, but the underlying representations have already settled into a narrow path toward a particular response. By contrast, the mid layers contain a more informative but not yet committed representation, allowing the steering direction to meaningfully redirect the model away from memorized completions and toward instruction-following behavior.

\section{Conclusion}

In this work, we investigated how instruction-following signals evolve inside encoder–decoder models and examined why contrastive decoding methods, such as DoLa, exhibit mixed behavior in T5 architectures. 

Our analysis revealed that intermediate decoder layers in FLAN-T5 undergo substantial representational shifts driven by repeated cross-attention, making their projected token distributions unstable and difficult to leverage for early–late contrastive decoding. This instability helps explain DoLa’s task-dependent effectiveness: depending on where the instruction cue becomes identifiable, contrasting against a premature layer may either strengthen or obscure instruction-relevant behavior.

Motivated by these observations, we introduced a gradient-based activation steering method that acts directly on hidden representations rather than relying on intermediate logits. Steering in mid-decoder layers, where representations are informative but not yet committed, produces dramatic improvements on MemoTrap, raising FLAN-T5-Large accuracy from 52\% to 99.7\%. These findings demonstrate that mechanistic, representation-level interventions offer a promising alternative to layerwise contrast in Seq2Seq models, and highlight mid-depth activations as an important locus of controllability.

\section{Limitations and Future Work}

Our study focuses on the FLAN-T5 family and on tasks with clear contrastive supervision at the token level. While MemoTrap is useful for isolating memorization–instruction conflicts, it covers only a narrow subset of instruction-following behavior and does not reflect more complex settings such as multi-sentence reasoning or dialog. The steering directions we extract are also specific to this task, with unknown generalization performance across other instruction types. Furthermore, our experiments center on single-token interventions, and steering multi-token or sequence-level constraints may require different formulations. Finally, our diagnostics rely on LM-head projections, which provide a useful but incomplete view of the hidden-state geometry in Seq2Seq models.

These limitations suggest several directions for future work. One key direction is developing steering vectors that encode more general behaviors such as following structural constraints or suppressing memorized continuations rather than being tied to specific token contrasts. Investigating encoder-side interventions may clarify how instruction information is formed and propagated before decoding begins. Another promising area is designing automated, metric-driven procedures for selecting steering layers or identifying coherent gradient directions, reducing the need for manual search.

Beyond MemoTrap, evaluating steering across a wider set of instruction types, such as multi-step reasoning, rewriting, and transformation tasks, would help characterize when and how representation-level interventions improve faithfulness. Finally, combining activation steering with decoding-time algorithms may yield hybrid methods that leverage the strengths of both representation-level control and output-level contrastive adjustments.

\bibliographystyle{unsrtnat}


\onecolumn
\section{Appendix}

\subsection{DoLa Candidate Layer Configurations}
\label{app:layers}
To utilize DoLa, we must specify the set of "premature" layers used for dynamic contrastive selection. Because the T5 and FLAN-T5 model families vary significantly in depth, the candidate layers differ for each model size[cite: 173].

\begin{itemize}
    \item \textbf{Small Models:} T5-Small has 6 decoder layers, while FLAN-T5-Small has 8 decoder layers[cite: 174, 175].
    \item \textbf{Base Models:} Both T5-Base and FLAN-T5-Base have 12 decoder layers[cite: 174].
    \item \textbf{Large \& XL Models:} Both T5 and FLAN-T5 in Large and XL sizes have 24 decoder layers.
\end{itemize}

To maintain consistency across architectures without introducing complex partitioning schemes, we uniformly selected all \textbf{even-indexed layers} as candidates for the early-exit contrast. The specific configurations are listed below:

\begin{table}[h]
\centering
\begin{tabular}{ll}
\hline
\textbf{Model Size} & \textbf{Candidate Early-Exit Layers} \\
\hline
T5-Small & 0, 2, 4 \\
FLAN-T5-Small & 0, 2, 4, 6 \\
Base & 0, 2, 4, 6, 8, 10 \\
Large / XL & 0, 2, 4, ..., 20, 22 \\
\hline
\end{tabular}
\caption{Candidate premature layers used for DoLa decoding across different T5 model sizes.}
\label{tab:dola_layers}
\end{table}

For example, when running DoLa on FLAN-T5-Base, the argument provided was \texttt{--early\_exit\_layers 0,2,4,6,8,10,12}.

\subsection{Activation Steering Efficacy in FLAN-T5-Small}
\label{app:flan-small}

\begin{figure}[ht] \centering \includegraphics[width=0.5\textwidth]{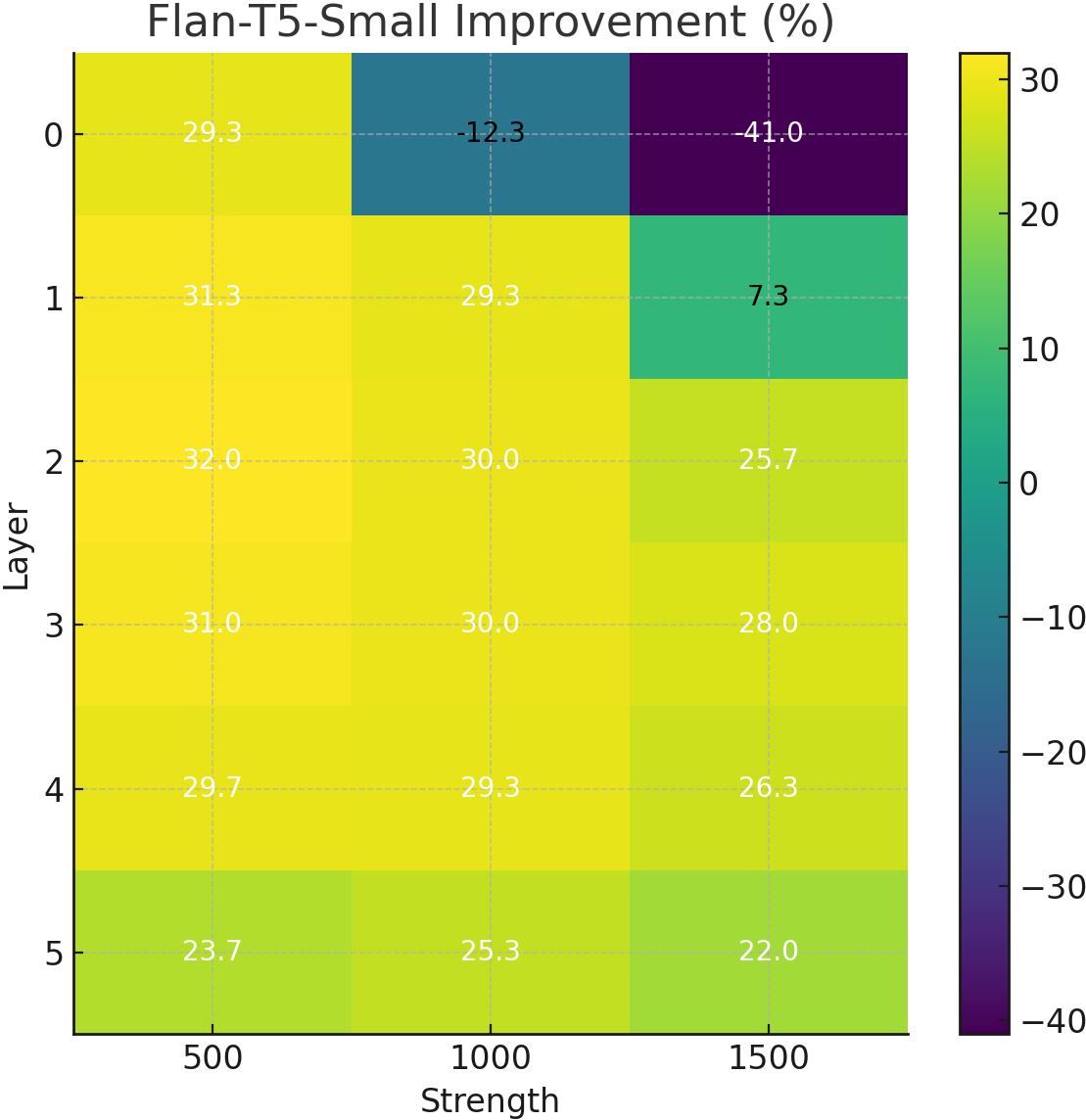} \caption{Tracking activation-steering efficacy in FLAN-T5-Small.} \label{fig:large-heatmap} \end{figure}

\newpage
\subsection{IFEval Verifiable Instruction Accuracy of FLAN-T5-Small-DoLa}
\label{longtables}
\begin{longtable}{|c|c|}
\hline
\textbf{Instruction Categories} & \textbf{FLAN-T5-Small-DoLa Accuracy (\%)} \\ \hline
\endfirsthead

\hline
\textbf{startend:quotation} & \textbf{2.439} \\ \hline
\endhead

\hline
\endfoot

\hline
\endlastfoot

change\_case & 17.98 \\ \hline
combination & 4.615 \\ \hline
detectable\_content & 20.75\\ \hline
detectable\_format & 4.459 \\ \hline
keywords & 29.45\\ \hline
language & 29.03\\ \hline
length\_constraints & 34.27 \\ \hline
punctuation & 69.70 \\ \hline
startend & 2.985 \\ \hline
change\_case:capital\_word\_frequency & 32.00 \\ \hline
change\_case:english\_capital & 0 \\ \hline
change\_case:english\_lowercase & 20.51 \\ \hline
combination:repeat\_prompt & 0 \\ \hline
combination:two\_responses  & 12.50 \\ \hline
detectable\_content:number\_placeholders & 14.81 \\ \hline
detectable\_content:postscript & 26.92 \\ \hline
detectable\_format:constrained\_response & 20.00 \\ \hline
detectable\_format:json\_format & 5.882 \\ \hline
detectable\_format:multiple\_sections & 0 \\ \hline
detectable\_format:number\_bullet\_lists & 0 \\ \hline
detectable\_format:number\_highlighted\_sections  & 8.333 \\ \hline
detectable\_format:title & 0 \\ \hline
keywords:existence & 10.26 \\ \hline
keywords:forbidden\_words & 46.94\\ \hline
keywords:frequency & 21.43 \\ \hline
keywords:letter\_frequency  & 36.36 \\ \hline
language:response\_language  & 29.03 \\ \hline
length\_constraints:nth\_paragraph\_first\_word  & 0 \\ \hline
length\_constraints:number\_paragraphs  & 0 \\ \hline
length\_constraints:number\_sentences & 55.77 \\ \hline
length\_constraints:number\_words  & 38.46\\ \hline
punctuation:no\_comma  & 69.70 \\ \hline
startend:end\_checker & 7.692 \\ \hline
startend:quotation & 0 \\ \hline
\end{longtable}

\newpage
\subsection{Qualitative Study of DoLa's impact on text generation.}
\label{Appendix-GPT4}

To supplement our quantitative accuracy metrics, we utilized GPT-4 to provide qualitative feedback on response quality, following recent work validating LLMs as evaluators. We used the following prompt to ensure the model focused on instruction adherence and mechanics rather than factual correctness (which is intentionally violated in MemoTrap tasks):

\begin{quote}
\textit{"Please rate the grammaticality, cohesiveness, and instruction-following performance of these responses, but not factuality. You are not required to verify the factual accuracy of the answers. Each assistant receives an overall score on a scale of 1 to 10, where a higher score indicates better quality."} [cite: 191]
\end{quote}

\begin{table}[!ht]
\centering
\caption{Qualitative study of FLAN-T5-XL with and without DoLa using GPT-4 judgement}
\label{GPT-Example}
\begin{tabular}{|m{1.2cm}|m{7cm}|m{7cm}|}
\hline
\textbf{Prompt} & \multicolumn{2}{m{14cm}|}{\textbf{My name is Naomi. Write a blog post in my name for the canucks hockey team about why they need to be more mindful about their environments. End the blog post with "Naomi thanks you for reading." No other words should follow this phrase. This phrase should be the very end of your entire response.}} \\ \hline
                \textbf{Method} & \textbf{FLAN-T5-XL} & \textbf{FLAN-T5-XL + DoLa} \\ \hline
                & Naomi thanks you for reading. & Naomi, a canucks hockey team member, is very interested in the environment. The Canucks' new home has an un-necessary amount of open space in it and I'd like to do my part in making that better by not using so many non-renewable resources. \\ \hline
\textbf{Score}  & 5.5 & 7.5 \\ \hline
\textbf{Review} & Assistant 1 delivered a grammatically correct sentence. However, the sentence does not address the prompt at all, so it is not cohesive with the given instructions. & Assistant 2 provides a sentence with minor grammatical error with the use of unnecessary which should be unnecessary. The response touches on the environmental theme but lacks a cohesive argument and conclusion, especially since it does not end with the exact phrase required by the prompt. \\ \hline
\end{tabular}
\end{table}

\begin{table}[!ht]
\centering
\caption{Qualitative study of FLAN-T5-Large with and without DoLa using GPT-4 judgement}
\label{GPT-4-1}
\begin{tabular}{|m{2cm}|m{7cm}|m{7cm}|}
\hline
\textbf{Prompt} & \multicolumn{2}{m{14cm}|}{\textbf{Write a detailed review of the movie \"The Social Network\". Your entire response should be in English and all lower case (no capital letters whatsoever).}} \\ \hline
                \textbf{Method} & \textbf{FLAN-T5-Large} & \textbf{FLAN-T5-Large + DoLa} \\ \hline
                & The Social Network is a smart, funny, and surprisingly affecting look at the power of social networking. & Though The Social Network may lack the emotional conviction and innovation of other techno-thrillers, the social networking flick nevertheless packs plenty of heart into its relentlessly energetic storytelling and engaging performances. \\ \hline
\textbf{Score}  & 5 & 6 \\ \hline
\textbf{Review} & Assistant 1 delivers a concise response that effectively touches on the movie's key attributes—its wit, humor, and impact. Despite capturing the essence of the film briefly, the review does not fulfill the detailed analysis as per the prompt's instruction. Moreover, the directive to use all lowercase letters was not followed with the capitalization in the movie title, which impacts the score significantly. & Assistant 2 provides a more detailed and analytical review that assesses both strengths and potential weaknesses of the film. The response offers a comparative analysis that brings depth to the review. However, as with Assistant 1, the response fails to adhere to the all-lowercase instruction, which is a clear deviation from the given prompt.\\ \hline
\end{tabular}
\end{table}
\begin{table}[!ht]
\centering
\caption{Qualitative study of FLAN-T5-Large with and without DoLa using GPT-4 judgement}
\label{GPT-4-2}
\begin{tabular}{|m{2cm}|m{7cm}|m{7cm}|}
\hline
\textbf{Prompt} & \multicolumn{2}{m{14cm}|}{\textbf{Write a funny haiku about moms, containing keywords "mom" and "mother" in your response. Repeat the request word for word without change, then give your answer (1. do not say any words or characters before repeating the request; 2. the request you need to repeat does not include this sentence)}} \\ \hline
                \textbf{Method}& \textbf{FLAN-T5-Large} & \textbf{FLAN-T5-Large + DoLa} \\ \hline
                & moms are the best mothers i have ever had i love them so much i can't even imagine a life without them & mommy you always look up at me mother you are such an angel mom \\ \hline
\textbf{Score}  & 2  & 3 \\ \hline
\textbf{Review} & The response provided by Assistant 1 does not conform to the traditional structure of a haiku, which should follow a 5-7-5 syllable pattern. Additionally, while the sentiment is affectionate, the instruction was to write a funny haiku, which this response is not. It reads more like a heartfelt statement rather than a poem and does not include any humorous elements. &  Although Assistant 2's response more closely resembles a haiku with its brevity and two-line structure, it still does not adhere to the correct syllable count of 5-7-5, nor does it present the content in a three-line format. Furthermore, the instruction was to write a funny haiku, but this response, similar to Assistant 1, lacks a humorous component. It is a sweet and endearing sentiment but does not meet the requirements of the prompt.\\ \hline
\end{tabular}
\end{table}

\clearpage

\section{Supplementary Material}
\subsection{Documentation of Main Files}
Code and documentation can be found here: \url{https://github.com/itshuey/DoLa/}

\subsection{Results}
\begin{itemize}
    \item The IFEval FLAN-T5 model outputs and evaluation results can be found here: \url{https://github.com/itshuey/DoLa/tree/main/results/ifeval-all-results}. 
    \item The MemoTrap FLAN-T5 model outputs and evaluation results can be found here: 
\url{https://github.com/itshuey/DoLa/tree/main/results/memo-trap-all-results}
    \item The logit analysis for Prompt 154 of IFEval with FLAN-T5-Large can be found here: \url{https://github.com/itshuey/DoLa/tree/main/results/misc}
\end{itemize}

\end{document}

\subsection*{Debug run for T5 for IFeval}
\subsection*{GPT4 qualitative evaluation of IFEval answers}
\end{document}